\documentclass[11pt]{article}
\usepackage{acl2018}
\usepackage{times}
\usepackage{url}

\usepackage{latexsym}
\usepackage{natbib}
\usepackage{xcolor}
\usepackage{float}
\usepackage{graphicx}
\usepackage{afterpage}

\def\Tref#1{Table #1}
\def\Fref#1{Figure #1}
\def\Sref#1{Section #1}

\def\clap#1{\hbox to 0pt{\hss #1\hss}}

\def\footurl#1{\footnote{\url{#1}}}

\aclfinalcopy

\begin{document}

\title{In Search for Linear Relations in Sentence Embedding Spaces}

\author{Petra Baran\v{c}\'{i}kov\'{a} and Ond\v{r}ej Bojar \\
  Charles University \\
  Faculty of Mathematics and Physics \\
  Institute of Formal and Applied Linguistics \\
  {\tt \{barancikova,bojar\}@ufal.mff.cuni.cz} \\
}  
\maketitle              % typeset the title of the contribution

\begin{abstract}

We present an introductory investigation into continuous-space vector 
representations of sentences. We acquire pairs of very similar sentences
differing only by a small alterations (such as change of a noun, adding 
an adjective, noun or punctuation) from datasets for natural language 
inference using a simple pattern method. We look into how such a small 
change within the sentence text affects its representation in the continuous 
space and how such alterations are reflected by some of the popular sentence 
embedding models. We found that vector differences of some embeddings
actually reflect small changes within a sentence.
\end{abstract}

\section{Introduction}

Continuous-space representations of sentences, so-called sentence embeddings,
are becoming an interesting object of study, consider e.g. the BlackBox workshop.\footurl{https://blackboxnlp.github.io/}
%\{tady bude odkaz na to zvlastni
%cislo JNLE, ktere jsem spoluredigoval}
Representing sentences in a continuous
space, i.e. commonly with a long vector of real numbers, can be useful in
multiple ways, analogous to continuous word representations (word embeddings). Word embeddings have
provably made downstream processing robust to unimportant input variations or
minor errors (sometimes incl. typos), they have greatly boosted the performance of many tasks
in low data conditions and can form the basis of empirically-driven lexicographic explanations of word meanings.

One notable observation was made in \cite{Mikolov13a}, showing that
several interesting relations between words have their immediate geometric
counterpart in the continuous vector space. 

\begin{figure}[t]
\caption{An illustration of a continuous multi-dimensional vector space representing individual sentences, a `space of sentences' (upper plot) where each sentence is represented as a dot. Pairs of related sentences are connected with arrows; dashing indicates various relation types. The lower plot illustrates a possible `space of operations' (here vector difference, so all arrows are simply moved to start at a common origin). The hope is that similar operations (e.g. all vector transformations extracted from sentence pairs differing in the speed of travel ``running instead of walking'') would be represented close to each other in the space of operations, i.e. form a more or less compact \emph{cluster}.}
\label{space-of-sentences}
\begin{center}
\includegraphics[width=\columnwidth]{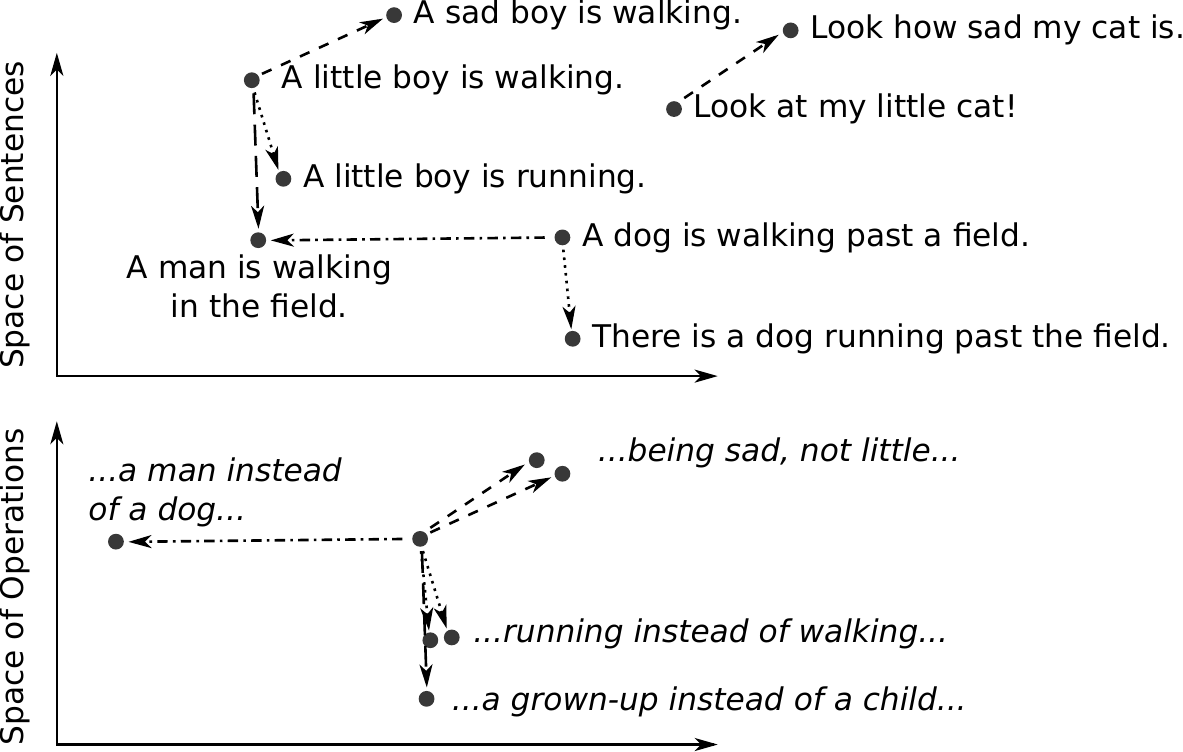}
\end{center}
\end{figure}

Our aim is to examine existing continuous representations of \emph{whole
sentences}, looking for an analogous behaviour. The idea of what we are hoping for is illustrated in
\Fref{space-of-sentences}. As with words, we would like to learn if and to what
extent some simple geometric
operations in the continuous space correspond to simple semantic operations on
the sentence strings. Similarly to \cite{Mikolov13a}, we are deliberately \emph{not including} this
aspect in the training objective of the sentence presentations but instead
search for properties that are learned in an unsupervised way, as a side-effect
of the original training objective, data and setup. This approach has the
potential of \emph{explaining} the good or bad performance of the examined types
of representations in various tasks.

%We present introductory investigation into a continuity of a multidimensional 
%vector space of sentences. In this research, we would like to look into 
%how even a small change (such a negation, change of tense, generalization) 
%in a sentence affects its representation and whether these changes are coherent 
%across a vector space or how to create a vector space with such a property.

The paper is structured as follows: \Sref{related} reviews the closest related work. \Sref{examined-sentences,examined-embeddings}, respectively, describe the dataset of 
sentences and the sentence embeddings methods we use. \Sref{operations} presents the
selection of operations on the sentence vectors. \Sref{experiments} provides 
the main experimental results of our work. We conclude in \Sref{conclusion}.

%However, obtaining such a data in good quality is time and money consuming. 
%Therefore, we started with a smaller scale experiment using existing data for 
%natural language inference (NLI). We selected slightly different sentences 
%using simple pattern method (see \cref{pattern_method}). The patterns are 
%actually not that common in data, but they represent a good example of 
%real-word sentences differing just by a small alternations.

\begin{figure*}[t]
\centering
\caption{Example of our pattern extraction method. In the first step, the longest common subsequence of tokens (\textit{ear is playing a guitar .}) is found and replaced with the variable X. 
In the second step, \textit{with a tattoo behind} is substituted with the variable Y. As the 
variables are not listed alphabetically in the premise, they are switched in the last step.}
\vspace*{2mm} 
\label{tab:howisitdone}
\resizebox{\linewidth}{!}{%
\begin{tabular}{ccc}
\clap{\textbf{step}}       & \textbf{premise}         & \textbf{hypothesis}    \\ \hline
\multicolumn{1}{|l|}{1.} & \multicolumn{1}{l|}{a man with a tattoo behind his \textbf{ear is playing a guitar .}} & \multicolumn{1}{l|}{a woman with a tattoo behind her\textbf{ ear is playing a guitar .}} \\
\multicolumn{1}{|l|}{2.} & \multicolumn{1}{l|}{a man \textbf{with a tattoo behind} his X}                        & \multicolumn{1}{l|}{a woman \textbf{with a tattoo behind} her X}                         \\
\multicolumn{1}{|l|}{3.} & \multicolumn{1}{l|}{a man Y his X}                                           & \multicolumn{1}{l|}{a woman Y her X}                                            \\
\multicolumn{1}{|l|}{4.} & \multicolumn{1}{l|}{a man X his Y}                                          & \multicolumn{1}{l|}{a woman X her Y}                                            \\ \hline
\end{tabular}
}
\end{figure*}

\begin{figure*}[t]
\caption{Top 10 patterns extracted from sentence pairs labelled as entailmens, contradictions and neutrals, respectively. Note the ``X $\to$ X" pattern indicating no change in the sentence string at all.}
\label{patterns}
\begin{center}
\resizebox{\linewidth}{!}{%
\begin{tabular}{rrlrrllrll}
  & \multicolumn{3}{c}{\textbf{entailments}}     & \multicolumn{3}{c}{\textbf{contradictions}}                & \multicolumn{3}{c}{\textbf{neutrals}}                       \\ \hline
\multicolumn{1}{|l|}{}    & premise   & hypothesis & \multicolumn{1}{l|}{}    & premise   & hypothesis & \multicolumn{1}{l|}{}    & premise    & hypothesis & \multicolumn{1}{l|}{}    \\ \hline
\multicolumn{1}{|l|}{1.}  & X         & X          & \multicolumn{1}{l|}{693} & X man Y   & X woman Y  & \multicolumn{1}{l|}{413} & X Y        & X sad Y    & \multicolumn{1}{l|}{701} \\
\multicolumn{1}{|l|}{2.}  & X man Y   & X person Y & \multicolumn{1}{l|}{224} & X woman Y & X man Y    & \multicolumn{1}{l|}{196} & X Y        & X big Y    & \multicolumn{1}{l|}{119} \\
\multicolumn{1}{|l|}{3.}  & X .       & X          & \multicolumn{1}{l|}{207} & X men     & X women Y  & \multicolumn{1}{l|}{111} & X Y        & X fat Y    & \multicolumn{1}{l|}{69}  \\
\multicolumn{1}{|l|}{4.}  & X woman Y & X person Y & \multicolumn{1}{l|}{118} & X boy Y   & X girl Y   & \multicolumn{1}{l|}{109} & X young Y  & X sad Y    & \multicolumn{1}{l|}{68}  \\
\multicolumn{1}{|l|}{5.}  & X boy Y   & X person Y & \multicolumn{1}{l|}{65}  & X dog Y   & X cat Y    & \multicolumn{1}{l|}{98}  & X people Y & X men Y    & \multicolumn{1}{l|}{60}  \\
\multicolumn{1}{|l|}{6.}  & X Y       & Y , X .    & \multicolumn{1}{l|}{61}  & X girl Y  & X boy Y    & \multicolumn{1}{l|}{97}  & X          & sad X      & \multicolumn{1}{l|}{51}  \\
\multicolumn{1}{|l|}{7.}  & X men Y   & X people Y & \multicolumn{1}{l|}{56}  & X women Y & X men Y    & \multicolumn{1}{l|}{64}  & X          & X          & \multicolumn{1}{l|}{41}  \\
\multicolumn{1}{|l|}{8.}  & two X     & X          & \multicolumn{1}{l|}{56}  & X Y,      & X not Y    & \multicolumn{1}{l|}{56}  & X person Y & X man Y    & \multicolumn{1}{l|}{34}  \\
\multicolumn{1}{|l|}{9.}  & X girl Y  & X person Y & \multicolumn{1}{l|}{55}  & two X,    & three X    & \multicolumn{1}{l|}{46}  & X Y        & X red Y    & \multicolumn{1}{l|}{30}  \\
\multicolumn{1}{|l|}{10.} & X , Y     & Y X .      & \multicolumn{1}{l|}{53}  & X child Y & X man Y    & \multicolumn{1}{l|}{44}  & X Y        & X busy Y   & \multicolumn{1}{l|}{28}  \\ \hline
\end{tabular}}
\end{center}
\end{figure*}

\section{Related Work}
\label{related}

Series of tests to measure how well their word embeddings capture semantic 
and syntactic information is defined in \cite{Mikolov13a}. These tests include 
for example declination of adjectives (``easy"$\rightarrow$``easier"$\rightarrow$``easiest"), 
changing the tense of a verb (``walking"$\rightarrow$``walk") or getting the capital (``Athens"$\rightarrow$``Greece") or currency of a state (``Angola"$\rightarrow$``kwanza").
\newcite{fasttext} and \newcite{subgram} have further refined the support of sub-word units, leading 
to considerable improvements in representing morpho-syntactic properties of words. 
 \newcite{VylomovaRCB15} largely extended the set of considered 
semantic  relations of words.

Sentence embeddings are most commonly evaluated extrinsically in so called `transfer tasks', 
i.e. comparing the evaluated representations based on their performance in sentence sentiment 
analysis, question type prediction, natural language inference and other assignments. \newcite{DBLP:journals/corr/abs-1805-01070} introduce `probing tasks' for intrinsic 
evaluation of sentence embeddings. They measure to what extent linguistic features like 
sentence length, word order, or the depth of the syntactic tree are available in a sentence 
embedding. This work was extended to SentEval \cite{senteval}, a toolkit for 
evaluating the quality of sentence embedding both intrinsically and extrinsically. It contains 
17 transfer tasks and 10  probing tasks. SentEval is applied to many recent sentence embedding 
techniques showing that no method had a consistently good performance across all tasks \cite{Perone2018}.

\newcite{worderror} examine how errors (such as incorrect word substitution caused by
automatic speech recognition) in a sentence affect its embedding. The embeddings of corrupted 
sentences are then used in textual similarity tasks and the performance is compared with 
original embedding. The results suggest that pretrained neural sentence encoders are much 
more robust to introduced errors contrary to bag-of-words embeddings.

\section{Examined Sentences}
\label{examined-sentences}

Because manual creation of sentence variations is costly, we reuse existing data from SNLI \citep{snli} and MultiNLI \citep{MultiNLI}.  Both these collections consist 
of pairs of sentences---a premise and a hypothesis---and their relationship (entailment/contradiction/neutral). 
The two datasets together contain 982k unique sentence pairs. All sentences were lowercased 
and tokenized using NLTK \citep{nltk}.

From all the available sentence pairs, we select only a subset where the difference between the sentences in the pair can be described with a simple pattern. Our method goes as follows: given two sentences, a premise $p$ 
and the corresponding hypothesis $h$, we find the longest common substring consisting of whole words and replace 
it with a variable. This is repeated once more, so our sentence patterns can have up to two variables. In the last 
step, we make sure the pattern is in a canonical form by switching the variables to ensure they are alphabetically sorted in $p$. 
The process is illustrated in \Fref{tab:howisitdone}.

Ten most common patterns for each NLI relation are shown in \Fref{patterns}. Many of the obtained patterns clearly match the sentence pair label. For instance the pattern no. 2 (``X man Y $\to$ X person Y'') can be expected to lead to a sentence pair illustrating entailment. If a man appears in a story, we can infer that a person appeared in the story. The contradictions illustrate typical oppositions like man--woman, dog--cat. Neutrals are various refinements of the content described by the sentences, probably in part due to the original instruction in SNLI that hypothesis ``might be a true'' given the premise in neutral relation.

We kept only 
patterns appearing with at least 20 different sentence pairs in order to have large 
and variable sets of sentence pairs in subsequent experiments. We also ignored the overall most common pattern, namely the identity, 
because it actually does not alter the sentence at all. Strangely enough, identity was observed not just among entailment pairs (693 cases), but also in neutral (41 cases) and contradiction (22) pairs.

Altogether, we collected 4,2k unique sentence pairs in 60 patterns. Only 10\% of this data comes from
MultiNLI, the majority is from SNLI.

\begin{table*}[]
\caption{This table presents the quality of pattern clustering in terms of the three cluster evaluation measures in the space of operations.
For all the scores, the value of 1 represents a perfect assignment and 0 corresponds
to random label assignment. All the numbers were computed using the Scikit-learn library \citep{scikit-learn}. 
Best operation according to each cluster score across the various embeddings in bold.}
\vspace*{2mm} 
\label{tab:vmeasure}
\resizebox{\linewidth}{!}{%
\begin{tabular}{llllllllllllll}
 &  & \multicolumn{4}{c}{Adjusted Rank Index} & \multicolumn{4}{c}{V-measure} & \multicolumn{4}{c}{Adjusted Mutual Information} \\ \hline
\multicolumn{1}{|l|}{embedding} & \multicolumn{1}{l|}{dim.} & - & + & * & \multicolumn{1}{l|}{/} & - & + & * & \multicolumn{1}{l|}{/} & - & + & * & \multicolumn{1}{l|}{/} \\ \hline
\multicolumn{1}{|l|}{\textbf{InferSent\_1}} & \multicolumn{1}{l|}{4096} & \textbf{0.58} & 0.03 & 0.03 & \multicolumn{1}{l|}{0.00} & \textbf{0.91} & 0.28 & 0.24 & \multicolumn{1}{l|}{0.03} & \textbf{0.87} & 0.18 & 0.14 & \multicolumn{1}{l|}{0.00} \\
\multicolumn{1}{|l|}{\textbf{ELMo}} & \multicolumn{1}{l|}{1024} & 0.55 & 0.03 & 0.02 & \multicolumn{1}{l|}{0.00} & 0.85 & 0.28 & 0.23 & \multicolumn{1}{l|}{0.03} & 0.82 & 0.18 & 0.13 & \multicolumn{1}{l|}{0.00} \\ 
\multicolumn{1}{|l|}{\textbf{LASER}} & \multicolumn{1}{l|}{1024} & 0.48 & 0.02 & 0.01 & \multicolumn{1}{l|}{0.00} & 0.79 & 0.19 & 0.15 & \multicolumn{1}{l|}{0.03} & 0.76 & 0.09 & 0.04 & \multicolumn{1}{l|}{0.00} \\
\multicolumn{1}{|l|}{\textbf{USE\_T}} & \multicolumn{1}{l|}{512} & 0.25 & 0.04 & 0.08 & \multicolumn{1}{l|}{0.00} & 0.73 & 0.25 & 0.30 & \multicolumn{1}{l|}{0.03} & 0.69 & 0.14 & 0.20 & \multicolumn{1}{l|}{0.00} \\
\multicolumn{1}{|l|}{\textbf{InferSent\_2}} & \multicolumn{1}{l|}{4096} & 0.31 & 0.04 & 0.04 & \multicolumn{1}{l|}{0.01} & 0.69 & 0.28 & 0.28 & \multicolumn{1}{l|}{0.10} & 0.65 & 0.19 & 0.19 & \multicolumn{1}{l|}{0.03} \\
\multicolumn{1}{|l|}{\textbf{BERT}} & \multicolumn{1}{l|}{1024} & 0.33 & 0.02 & 0.01 & \multicolumn{1}{l|}{0.00} & 0.66 & 0.22 & 0.16 & \multicolumn{1}{l|}{0.03} & 0.62 & 0.12 & 0.06 & \multicolumn{1}{l|}{0.00} \\
\multicolumn{1}{|l|}{\textbf{USE\_D}} & \multicolumn{1}{l|}{512} & 0.21 & 0.05 & 0.08 & \multicolumn{1}{l|}{0.00} & 0.65 & 0.27 & 0.33 & \multicolumn{1}{l|}{0.03} & 0.58 & 0.17 & 0.23 & \multicolumn{1}{l|}{0.00} \\
\hline
\multicolumn{1}{|l|}{average} & \multicolumn{1}{l|}{1775} & \textbf{0.39} & 0.03 & 0.04 & \multicolumn{1}{l|}{0.00} & \textbf{0.75} & 0.25 & 0.24 & \multicolumn{1}{l|}{0.04} & \textbf{0.71} & 0.15 & 0.14 & \multicolumn{1}{l|}{0.00} \\ \hline
\end{tabular}}
\end{table*}

\section{Sentence Embeddings}
\label{examined-embeddings}
We experiment with several popular pretrained sentence embeddings.

InferSent\footurl{https://github.com/facebookresearch/InferSent} \citep{infersent} 
is the first embedding model that used a supervised learning to compute sentence representations. It was 
trained to predict inference labels on the SNLI dataset. The authors 
tested 7 different architectures and BiLSTM encoder with max pooling achieved 
the best results. InferSent comes 
in two versions: \textbf{InferSent\_1} is trained with Glove embeddings 
\citep{glove} and \textbf{InferSent\_2} with fastText \citep{fasttext}.
InferSent representations are by far the largest, with the dimensionality of 4096 in both versions.

Similarly to InferSent, Universal Sentence Encoder \cite{use} uses unsupervised 
learning augmented with training on supervised data from SNLI. There are two models 
available. \textbf{USE\_T}\footurl{https://tfhub.dev/google/universal-sentence-encoder-large/3}
is a transformer-network \citep{Vaswani2017AttentionIA} designed for higher accuracy at
the cost of larger memory use and computational time.  \textbf{USE\_D}\footurl{https://tfhub.dev/google/universal-sentence-encoder/2} is a deep averaging network \cite{iyyer2015}, 
where words and bi-grams embeddings are averaged and used as input to a deep neural network 
that computes the final sentence embeddings. This second model is faster and more efficient
but its accuracy is lower. Both models output representation with 512 dimensions.

Unlike the previous models, \textbf{BERT}\footurl{https://github.com/google-research/bert} 
(Bidirectional Encoder Representations from Transformers) \cite{bert} is a deep unsupervised 
language representation, pre-trained using only unlabeled text. It has two self-supervised 
training objectives - masked language modelling and next sentence classification. It is 
considered bidirectional as the Transformer encoder reads the entire sequence of words at 
once. We use a pre-trained BERT-Large model with Whole Word Masking. BERT gives embeddings 
for every (sub)word unit, we take as a sentence embedding a [CLS] token, which is inserted 
at the beginning of every sentence. BERT embeddings have 1,024-dimensions.

\textbf{ELMo}\footurl{https://github.com/HIT-SCIR/ELMoForManyLangs}
(Embedding from Language Models) \cite{elmo} uses representations from a biLSTM that is trained 
with the language model objective on a large text dataset. Its embeddings are a function 
of the internal layers of the bi-directional Language Model (biLM), which should capture not 
only semantics and syntax, but also different meanings a word can represent in different contexts 
(polysemy). Similarly to BERT, each token representation of ELMo is a function of the entire 
input sentence - one word gets different embeddings in different contexts. ELMo computes an 
embedding for every token and we compute the final sentence embedding as the average over all 
tokens. It has dimensionality 1024.

\textbf{LASER}\footurl{https://github.com/facebookresearch/LASER} (Language-Agnostic 
SEntence Representations) \cite{laser} is a five-layer bi-directional 
LSTM (BiLSTM) network. The 1,024-dimension vectors are obtained by max-pooling over its last states. 
It was trained to translate from more than 90 languages to English or Spanish
at the same time, the source language was selected randomly in each batch.

\section{Choosing Vector Operations}
\label{operations}
\newcite{Mikolov13a} used a simple vector difference as the operation 
that relates two word embeddings. For sentences embeddings, we experiment a little 
and consider four simple operations: addition, subtraction, multiplication and 
division, all applied elementwise.
More operations could be also considered as long as they are reversible, so that we 
can isolate the vector change for a particular sentence alternation and apply it to 
the embedding of any other sentence. Hopefully, we would then land in the area where 
the correspondingly altered sentence is embedded.
%transformace: \url{https://en.wikipedia.org/wiki/Transformation_matrix} nebo ilustrovanejsi \url{https://www.continuummechanics.org/coordxforms.html}.}

The underlying idea of our analysis was already sketched in \Fref{space-of-sentences}.
From every sentence pair in our dataset, we extract the pattern, i.e. the string edit 
of the sentences. The arithmetic operation needed to move from the embedding of the 
first sentence to the embedding of the second sentence (in the continuous space of 
sentences) can be represented as a point in what we call the \emph{space of operations}. 
Considering all sentence pairs that share the same edit pattern, we obtain many points 
in the space of operations. If the space of sentences reflects the particular edit 
pattern in an accessible way, all the corresponding points in the space of operations 
will be close together, forming a cluster.

To select which of the arithmetic operations best suits the data, we test pattern clustering with three
common clustering performance evaluation methods:

\begin{itemize}
    \item \textbf{Adjusted Rand index} \cite{Hubert1985} is measure of the similarity between 
    two cluster assignments adjusted with chance normalization. The score ranges from −1 to +1 with 
    1 being the perfect match score and values around 0 meaning random label assignment. 
    Negative numbers show worse agreement than what is expected from a random result.
    \item \textbf{V-measure} \cite{vmeasure} is harmonic mean of \textit{homogeneity} (each 
    cluster should contain only members of one class) and \textit{completeness} (all members 
    of one class should be assigned to the same cluster). The score ranges from 0 (the worst
    situation) to 1 (perfect score). 
    \item \textbf{Adjusted Mutual Information} \cite{Strehl:2002} measures the agreement of 
    the two clusterings with the correction of agreement by chance. The random label
    assignment gets a score close to 0, while two identical clusterings get the score of 1. 
\end{itemize}

As the detailed description of these measures is out of scope of this article, we refer readers
to related literature (e.g. \cite{5211310}). We use these scores to compare patterns 
with labels predicted by k-Means (best result of 100 random initialisations). 
The results are presented in \Tref{tab:vmeasure}. It is apparent that the best distribution 
by far is achieved using the most intuitive operation, vector subtraction. 

There seems to be a weak correlation between the size of embeddings and the scores. The smallest 
embeddings USE\_D and USE\_T are getting the worst scores, while the largest embeddings InferSent\_1 
are the best scoring embeddings. However, InferSent\_2 with dimensionality 4096 is performing
poorly. The fact that several of the embeddings were trained on SNLI does not to seem benefit 
those embeddings. Between the three top scored embeddings, only InferSent\_1 was trained on the data 
that we use for evaluation of embeddings.

\section{Experiments}
\label{experiments}

For the following exploration of the continuous space of operations, we focus only on the ELMo embeddings. 
They scored second best in all scores but unlike the best scoring Infersent\_1, ELMo was not trained on SNLI, 
which is the major source of our sentence pairs.

\begin{figure*}[tb]
\caption{t-SNE representation of patterns. The points in the operation space are obtained by subtracting the
ELMo embedding of the hypothesis from the ELMo embedding of the premise. Best viewed in color. Colors correspond to the sentence patterns.}
\vspace*{15mm} 
\label{tsne}
\includegraphics[width=\textwidth]{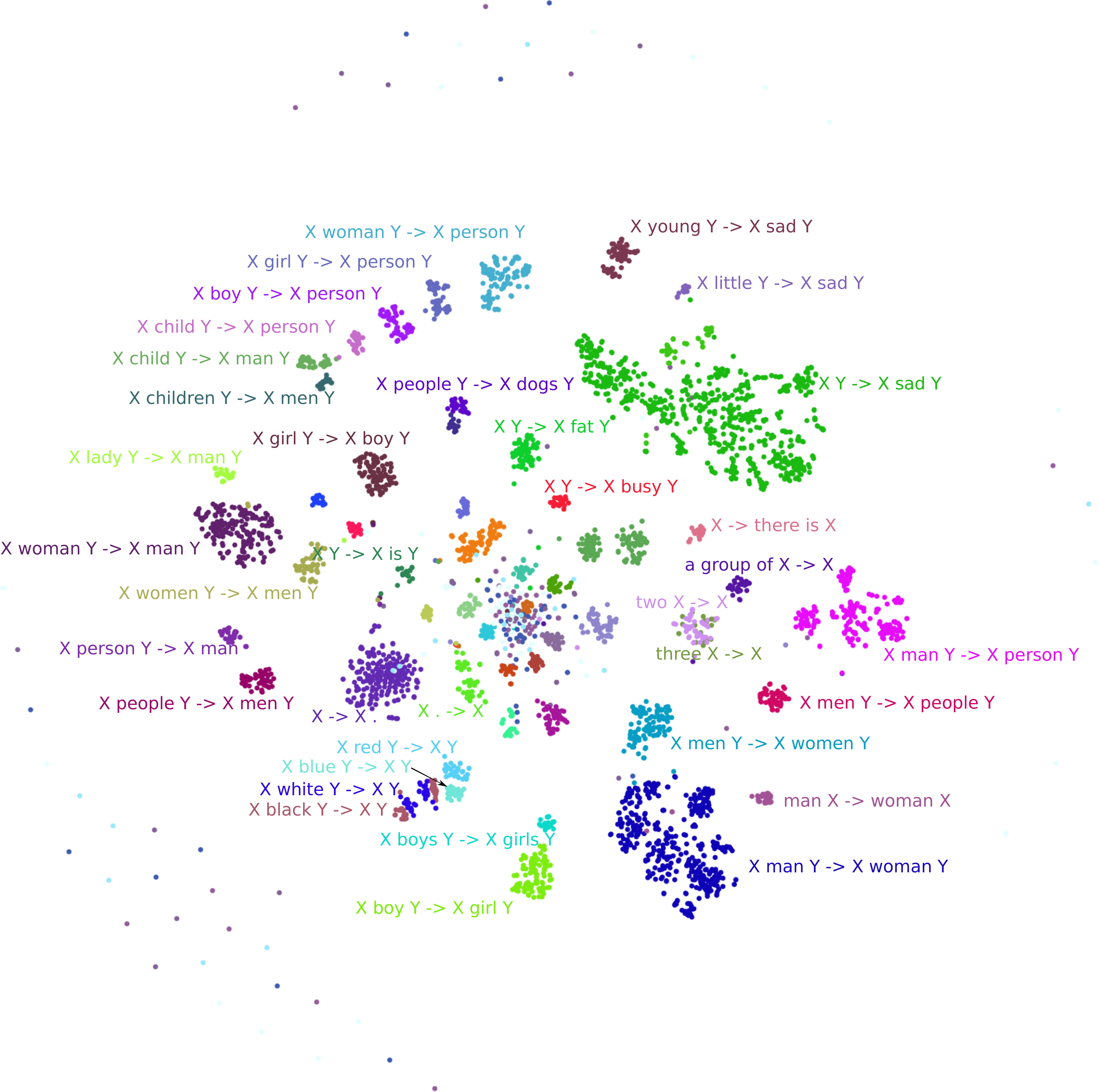}
\end{figure*}

The t-SNE \citep{vanDerMaaten2008} visualisation of subtractions of ELMo vectors is presented in \Fref{tsne}. 
The visualisation is constructed automatically and, of course, \emph{without} the knowledge of the pattern label. It shows 
that the patterns are generally grouped together into compact clusters with the exception
of a `chaos cloud' in the middle and several outliers.  Also there are several patterns that seem inseparable, e.g. ``two X $\to$ X" and ``three X $\to$ X", or ``X white Y $\to$ X Y" and 
``X black Y $\to$ X Y". 

We identified the patterns responsible for the noisy center and outliers by computing 
weighted inertia for each pattern (the sum of squared distances of samples to their cluster 
center divided by the size of sample). The clusters with highest inertia consists of 
patterns representing a change of word order and/or adding or removing punctuation. 
These patterns are:

\begin{table}[H]
\begin{tabular}{l|l|l}
X is Y . $\to$ Y is X & X Y . $\to$ Y X . & X $\to$ X . \\
X , Y . $\to$ Y X .  & X , Y . $\to$ Y , X . & \\
X Y . $\to$ Y , X . & X . $\to$ X & \\
\end{tabular}
\end{table}

%\XXX{Ja tomu sdeleni tady vubec nerozumim, mj. nevim z hlavy, ktere embeddingy jsou BoW.}
%It is apparent, the changes of word order is especially hard for BOW embedding \todo{tu to formulovat nejak smysluplne plus to, ze na obrazku je pridavani a odebravani tecky relativne peknej clustrik, takze se to asi tyka jenom carky}.

\afterpage{\clearpage}

\begin{figure*}[tb]
\caption{t-SNE representation of patterns as in \Fref{tsne} with colors coding now fully automatic clusters. Each cluster is labelled with the set of patterns extracted from sentence pairs assigned to the cluster. The numbers in parentheses indicate how many sentence pairs belong to the given pattern within this cluster and overall, resp. For instance the line ``two X $\rightarrow$ X (52/56)'' says that of the 56 sentence pairs differing in the prefix ``two'', 52 were automatically clustered together based on the subtraction of their ELMo embeddings.}
\vspace*{15mm} 
\label{clusters}
\includegraphics[width=\textwidth]{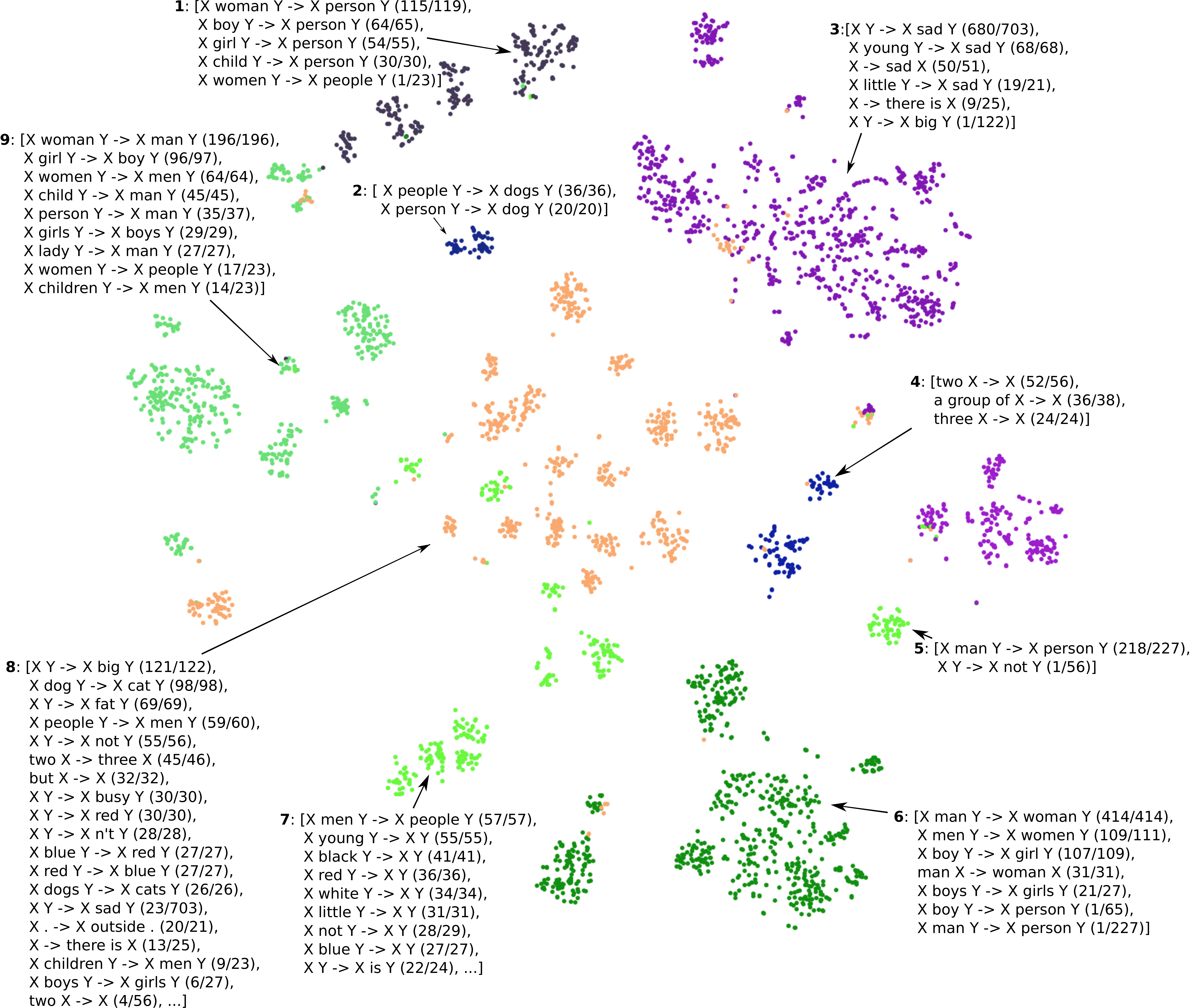}
\end{figure*}

To see if the space of operations can be interpreted also automatically, i.e. if the sentence relations are generalizable,
%how well are the patterns generalizable, i.e. if there are well defined broader, hypernymous patter in the vector space,
we remove the noisy patterns as above and apply fully unsupervised clustering: we do not even disclose the expected number of patterns, i.e. clusters.
We try two metrics for finding the optimal number of clusters: Davies-Bouldin's index \citep{Davies} and 
Silhouette Coefficient \citep{Rousseeuw:1987}. They are both designed to measure compactness and separation of the clusters, i.e. they award dense clusters that are far from each other. Both Davies-Bouldin index and Silhouette Coefficient agree that the best separation is achieved at 9 clusters. Running k-Means with 9 clusters, we get the result as plotted in \Fref{clusters}.

Manually inspecting the contents of the automatically identified clusters, we see that many clusters are meaningful in some way. For instance, Cluster 1 captures 90\% (altogether 264 out of 292) sentence pairs exerting the pattern of generalizing women, boys or girls to people. The counterpart for men belonging to people is spread into Cluster 5 (218 out of 227 pairs) for the singular case and not so clean Cluster 7 containing 57/57 of the plural pairs ``X men Y $\rightarrow$ X people Y'' together with various oppositions.
Cluster 2 covers all sentence pairs where a person is replaced with a dog.
Cluster 3 is primarily connected with sentence pairs introducing bad mood. 
Cluster 4 unites patterns that represent omitting a numeral/group.
Cluster 6 covers gender oppositions in one direction and Cluster 9 adds the other direction (with some noise for child/man and person/man and similar), etc.

\section{Conclusion and Future Work}
\label{conclusion}

We examined vector spaces of sentence representations as inferred automatically by sentence 
embedding methods such as InferSent or ELMo. Our goal was to find out if some simple arithmetic 
operations in the vector space correspond to meaningful edit operations on the sentence strings.

Our first explorations of 60 sentence edit patterns document that this is indeed the case. 
Automatically identified frequent patterns with 20 or more occurrences in the SNLI and MultiNLI 
datasets correspond to simple vector differences. The ELMo space (and others such as Infersent\_1, 
LASER and USE-T, which are omitted due to paper length requirements) exerts this property very well. 

Unfortunately, choosing ELMo as example might not have been the best option -- we compute ELMo 
embeddings by averaging contextualized word embeddings and majority of the patterns are just 
removing/adding/changing a single word. Difference between two such sentence embeddings may be a 
simple difference between the embeddings of the words substituted, depending on the effect of the contextualization. Thus, the differences in vector space would show rather the word embeddings 
than the sentence embeddings. 

%we plan to focus on more creative sentence variations.
       
It should be noted that our search made use of only about 0.5\% of the sentence pairs available in SNLI and MultiNLI. The remaining sentence pairs differ beyond what was extractable automatically using our simple pattern method. A different approach for a fine-grained description of the semantic relation between two sentences would have to be taken for a better exploitation of the available data.

Our plans for the long term are to further verify these observations using a more diverse set of vector operations and a larger set of sentence alternations, primarily by extending the set of alternation \emph{types}. We also plan to examine the possibilities of \emph{generating sentence strings} back from the sentence embedding space. If successful, our method could lead to controlled paraphrasing via the continuous space: take an input sentence, embed it, modify the embedding using a vector operation and generate the target sentence in the standard textual from.

\section*{Acknowledgment}
This work has been supported by the grant No. 18-24210S of the Czech Science Foundation. It has been using language resources and tools stored and distributed by the LINDAT/CLARIN project of the Ministry of Education, Youth and Sports of the Czech Republic (project LM2015071).

\bibliographystyle{acl_natbib}
\bibliography{biblio}
\end{document}